\documentclass{article}
\usepackage{spconf,amsmath,graphicx,hyperref}

\usepackage{multirow}
\usepackage{array}
\usepackage{xspace}
\usepackage{booktabs}
\usepackage{siunitx}
\usepackage{algorithm}
\usepackage{algpseudocode}
\usepackage{amssymb}   % habilita \checkmark
\usepackage{xcolor}
\usepackage{enumitem} % enables [noitemsep] in description lists
\definecolor{prompt}{HTML}{6B7280}     % gray-500
\definecolor{speech}{HTML}{2563EB}     % blue-600  (projector outputs)
\definecolor{noiseA}{HTML}{0D9488}     % teal-600  (nearest-token replacements)
\definecolor{noiseT}{HTML}{DB2777}     % pink-600  (distorted text)
\definecolor{textout}{HTML}{EA580C}    % orange-600 (target transcript)
\definecolor{softp}{HTML}{8B5CF6}      % violet-600 (soft prompts)
\ninept 
\title{Slot Filling as a Reasoning Task for SpeechLLMs}
\name{
 \begin{tabular}{c}
 Kadri Hacioglu, Manjunath K E, Andreas Stolcke
 \end{tabular}
}
\address{Uniphore \quad
}
\begin{document}
%\ninept
%
\maketitle
\begin{abstract} 
We propose integration of reasoning into speech large language models (speechLLMs) for the end-to-end slot-filling task. Inspired by the recent development of reasoning LLMs, we use a chain-of-thought framework to decompose the slot-filling task into multiple reasoning steps, create a reasoning dataset and apply the supervised fine-tuning strategy to a speechLLM. We distinguish between regular and reasoning speechLLMs and experiment with different types and sizes of LLMs as their text foundation models.  We demonstrate performance improvements by introducing reasoning (intermediate) steps. However, we show that a reasoning textual LLM developed mainly for math, logic and coding domains might be inferior as a foundation model for a reasoning speechLLM. We further show that hybrid speechLLMs, built on a hybrid text foundation LLM and fine-tuned to preserve both direct and reasoning modes of operation, have better performance than those fine-tuned employing only one mode of operation.
\end{abstract}
\section{Introduction}
\label{sec:intro}
\begin{figure}[t]
  \centering
  \vspace{-15pt}
  \includegraphics[scale=0.22]{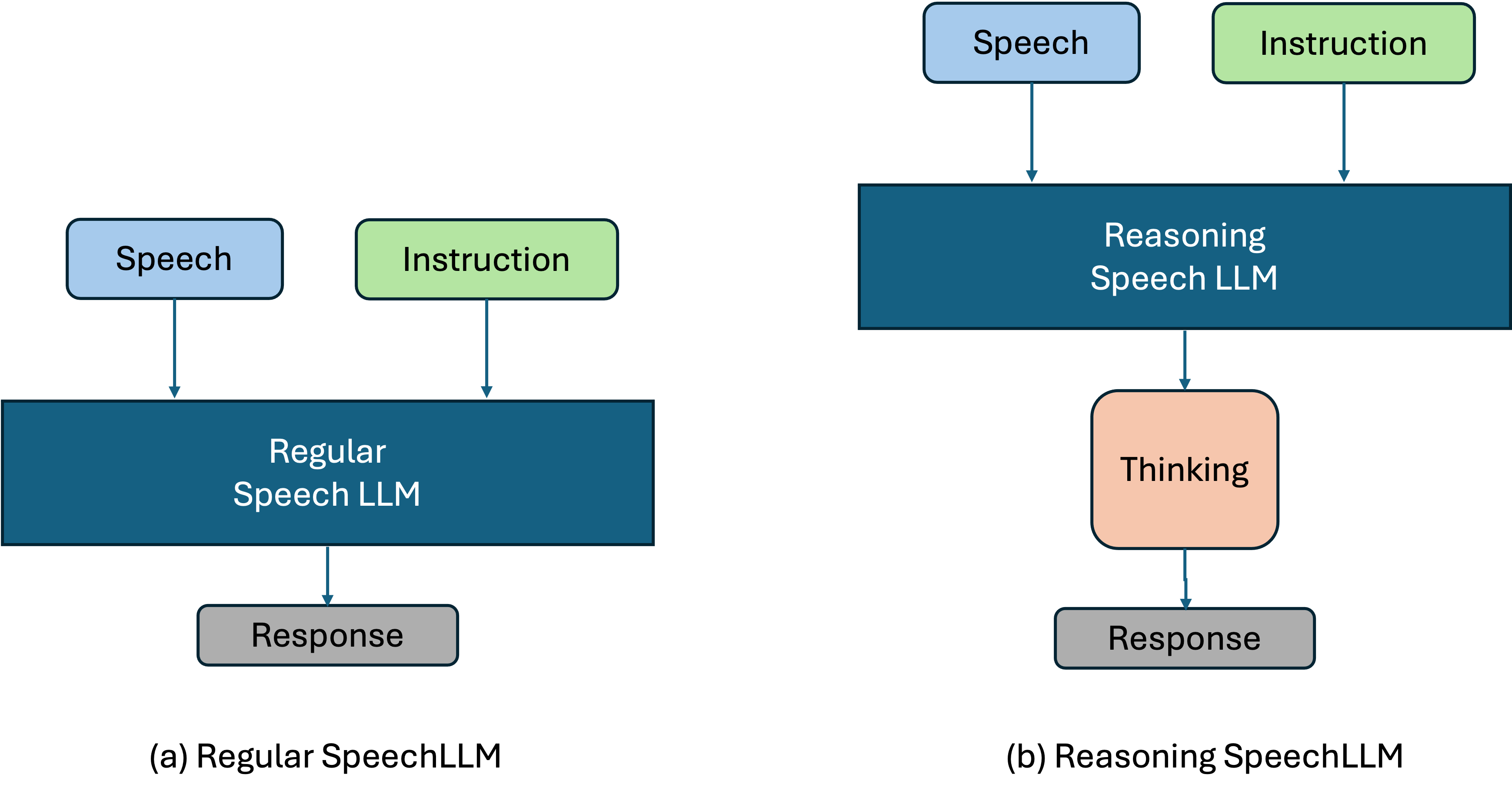}
  \vskip -0.5cm
  \caption{SpeechLLM types}
  \label{fig:types}
  \vspace{-15pt}
\end{figure}

Emerging reasoning large language models (LLMs), such as Openai-o1 \cite{b1} and Deepseek-r1 \cite{b2}
 are typically used for tasks requiring multi-step problem solving, logical inference, and mathematical reasoning while “thinking” prior to providing their final answer. In contrast, regular (non-reasoning) LLMs, such as closed-source GPT-4 \cite{b4} and open-source LLama3 series \cite{b5} focus on direct response generation without explicit chain-of-thought (CoT) reasoning traces. Prior work has demonstrated that reasoning capabilities in LLMs can be significantly enhanced by techniques such as CoT prompting \cite{b6,b7}, supervised fine tuning (SFT) and reinforcement learning (RL) \cite{b8,b10}. In this work we adopt SFT as our primary method. With the rise of multimodal LLMs, in particular speechLLMs \cite{b11, b12}, spoken language understanding tasks are being addressed in a unified, end-to-end, generative, and instruction following manner, in either a single-task or a multi-task setting. 
\begin{figure}[t]
\vspace{-10pt}
  \centering
  \includegraphics[scale=0.22]{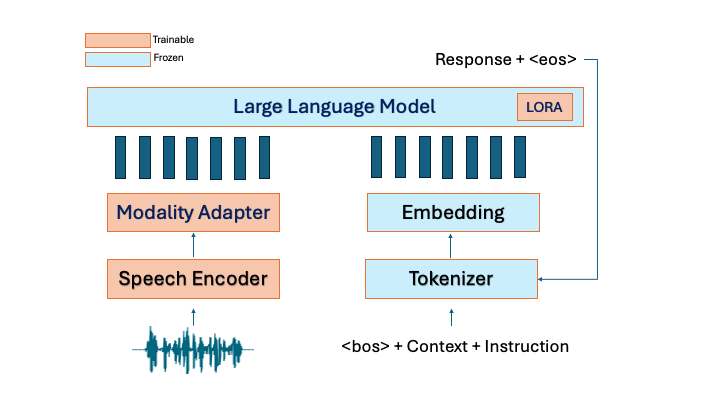}
  \vskip -0.5cm
  \caption{SpeechLLM architecture}
  \label{fig:architecture}
\end{figure}
Here we introduce a framework using CoT reasoning for slot filling, thereby transforming the single-step approaches \cite{b13} into a multi-step approach.
By constructing a dataset with intermediate reasoning steps, we enable speechLLMs to learn multi-step slot filling through supervised fine-tuning. Among many possible speechLLM architectures \cite{b14,b15,b17,b18,b19,b20,b21,b22,b24}, we considered and implemented one that has a structure very similar to SpeechVerse \cite{b25}.
As shown in Figure~\ref{fig:architecture}, it consists of three main components: a speech encoder, a modality adapter, and a pretrained large language model (LLM). This architecture is designed to jointly process speech signals and textual instructions, enabling it to perform a wide range of tasks that require understanding of both speech and text. The foundation models, pretrained at scale in their respective modalities, promise data-efficient development. Moreover, small-scale modality adapters for alignment and large-scale foundation models with parameter-efficient fine-tuning enable computationally efficient development. Importantly, the emergent abilities of LLMs, if preserved during fine-tuning, offer the potential for instruction-based zero-shot learning and improved generalization. Based on the nature of inputs and responses, we categorize speechLLMs into two broad types as shown in Figure~\ref{fig:types}: (a) regular (non-reasoning) speechLLM and (b) reasoning speechLLM. The former lacks COT traces at their outputs where the latter has its thought traces between special tags for "thinking" before they provide their final answer enclosed between special tags for "response". In our study, we aim to explore whether a reasoning-capable LLM is useful, particularly, one that can emulate the multi-step thought processes humans might use when approaching the task. Traditionally, slot filling has been treated as a single-step prediction problem, but we investigate whether reframing it as a reasoning task might yield better performance. This brings us to key questions: What type of foundation LLM is most appropriate for this task: a base model, an instruction-tuned model, or one optimized for reasoning? And to what extent is model scale a prerequisite for reasoning capabilities to emerge? While larger models may offer improved reasoning performance, they come with increased cost, so we aim to understand whether smaller, more efficient models can achieve similar reasoning behaviors when fine-tuned effectively. Through this exploration, we seek to balance task formulation, model type, and compute efficiency to determine the most effective approach for slot filling. Our key contributions are:
\begin{itemize}
  \item A new formulation of slot filling as a reasoning task, mimicking a plausible multi-step human annotation workflow.
  \item Implementation of this formulation using speechLLMs via supervised fine-tuning to  generate reasoning traces followed by structured slot/value responses.
  \item A comprehensive analysis of the speechLLM performance across a diverse set of text foundation LLMs,  varying in size and reasoning ability.
  \item A new hybrid speechLLM fine-tuning method including both direct and reasoning-style supervision for better and balanced performance.
  \item Empirical insights to guide development of speechLLMs for speech understanding tasks.
\end{itemize}

\section{Data Preparation}
\label{sec:annotation}
\subsection{Description and Annotation }

The data for slot filling is a set of scripted call center conversations between agents and customers curated by DefinedAI\footnote{https://www.defined.ai}. This collection has approximately 31K calls with almost 1M turns and 2.1K hours of speech. The domains covered are banking, telecommunication, insurance and retail. We used GPT-4o to annotate our dataset with slot labels and values. To achieve broad, diverse and open-ended slot filling, we decided not to prime the LLM for any specific set of slot labels. Instead, given the entire call itself, we instructed the LLM to do slot filling turn-by-turn for mentions that reflect real world entities, events, dates, times and  numerals while avoiding abstract notions of “entities” such as issues, solutions, broader concepts, advice or ideas. 
%The structure of the prompt for GPT-4o to annotate each turn in a complete conversation sample between an agent and a user is shown in Figure~\ref{fig:gpt-prompt}. Further insights into our dataset construction, annotation, human validation and s are provided in \cite{26} which covers a related task of slot-filling and a subset of the dataset used.   

\subsection{Instruction-based Dataset}
\label{sec:instruction}

We now describe the conversion of the slot filling dataset into an instruction-based training dataset. This dataset consists of three fields: audio, instruction, and desired output. Instructions are the most crucial part of the data preparation. It should consist of a description of the task in natural language. Although it is possible to use a fixed instruction, a diverse set of instructions is beneficial to  generalize to novel prompts not seen in training. In addition, we introduced several strategies for improving coverage across a variety of use-cases of slot filling. We assume turn-by-turn slot filling with/without context, and with/without querying specific slots. The context is defined as the recognition results for the previous $T$ turns. We randomize the context size in the range $0 \le T \le 3$. When we query specific slots in the prompt, we take the ground truth slots in the corresponding turn, if any, and then add a varying numbers $S$ of distractors. We also randomize the number of distractors, $1 \le S \le 5$. For each case, we randomly sample from a set of 10 prompts. 
%In Figure~\ref{fig:task-prompts}, a subset of instructions used for the training data of the speechLLM model is shown.

\subsection{Reasoning Dataset}
 As introduced earlier, in standard slot filling datasets, each speech utterance is paired with a structured JSON format containing slot types and their corresponding values, typically generated from direct text-to-label mappings. However, this direct  and single step mapping bypasses the implicit cognitive steps a human annotator takes. Therefore, to convert these regular slot filling datasets into reasoning-style datasets, we follow a process inspired by how humans would naturally perform the annotation task.  

Human listeners, much less annotators, do not necessarily assign slot labels directly upon hearing speech. Instead, they may first listen to the utterance, then mentally transcribe it into text using their language and world knowledge. After obtaining the transcript, they may then interpret its meaning and identify relevant spans or mentions of real world object types. Finally, following annotation guidelines, they assign appropriate slot labels and populate them with values, if present, based on contextual reasoning.  

To reflect this process, we transform regular slot filling data into multi-step reasoning-style data. Each example is augmented with intermediate reasoning steps, including (1) transcription of the speech, (2) identification of potential spans or phrases associated with each slot type, and (3) justification for assigning each slot-value pair. This COT format allows models to simulate the human annotation workflow, encouraging deeper understanding and interpretability. By structuring the task this way, we shift from single-step prediction to a multi-step reasoning task, laying the foundation for training and evaluating reasoning-capable Speech LLMs for slot filling. The examples for regular and reasoning outputs are shown in Figure~\ref{fig:reasoning_output}.
\begin{figure}[t]
\centering
\vspace{-10pt}
\begin{minipage}{0.95\columnwidth}
\footnotesize
\fbox{%
\begin{tabular}{p{0.95\columnwidth}}
\small
\textbf{Prompt}: $<$audio$>$$<$bos$>$Find slot values for new$\_$limit, family$\_$members$\_$count, review$\_$period, payment$\_$frequency, payment$\_$amount in the current audio. Format the output as JSON.\\

\textbf{Regular response}: \\
\{'payment$\_$frequency': 'monthly', 'payment$\_$amount': '€30', 'new$\_$limit': 'None', 'family$\_$members$\_$count': 'None', 'review$\_$period': 'None'\}$<$\textbackslash eos$>$ \\
\textbf{Reasoning response}: \\
$<$thinking$>$\\
I hear the utterance in the audio clip is \\``` Ok, thanks again for calling today, \"Patrick\". And you are paying a month- you have a monthly payment set up for €30 a month. Is that correct? ```\\I see that the information bearing mentions in the utterance are monthly $|$ €30.\\The labels queried for are payment$\_$frequency, payment$\_$amount, new$\_$limit, family$\_$members$\_$count, review$\_$period\\Based on the semantics of payment$\_$frequency, payment$\_$amount slots, the mentions in the utterance can be assigned to them. The others are all 'None'\\$<$\textbackslash thinking$>$\\ $<$response$>$\\\{'payment$\_$frequency': 'monthly', 'payment$\_$amount': '€30', 'new$\_$limit': 'None', 'family$\_$members$\_$count': 'None', 'review$\_$period': 'None'\}\\ $<$\textbackslash response$>$$<$\textbackslash eos$>$
\end{tabular}
}

\caption{ Comparison of regular and reasoning-style slot-filling responses. }
\label{fig:reasoning_output}
\end{minipage}
\vspace{0pt}
\end{figure}

%, This paper presents a proposal that avoids this by combining multiple data views during fine-tuning, alternating between (i) projector outputs derived from audio, (ii) nearest-token approximations of those outputs, and (iii) synthetically distorted text.

%The result is a target-audio–free recipe that maintains speech-processing capacity and improves domain-specific performance.
% \section*{Acknowledgments}
% This work was supported by the Idiap Research Institute and Uniphore collaboration project.

%\small
%%
\section{Experiments}
\subsection{Experimental Setup}
Experiments were conducted using 4x10G GPUs.
We employed PEFT  using QLoRA. LORA settings were selected as $rank=32$,  $\alpha=128$ and $dropout=0.05$ for all linear target modules. The batch size was 4 per GPU, with gradient accumulation of 8 steps, resulting in an effective batch size of 128. A cosine scheduler, over 10-15 epochs, was used with a maximum learning rate of $2 \times 10^{-4}$. We implemented a linear warm-up for the first 20\% of the total number of iterations. For optimization, we employed the AdamW (weighted Adam) optimizer. We applied gradient clipping with 1.0 threshold. 

\subsection{Text Foundation LLMs}
We construct several speechLLM systems that differ only in the underlying text foundation LLM, while keeping the speech encoder and modality adapter the same size and type. We use the encoder of Whisper-base \cite{b27} and introduce a lightweight modality adapter consisting of a frame-stacked two-layer MLP that performs temporal stacking by a stack factor of 4 frames over the already 2x downsampled output of Whisper encoder, resulting in an effective 8x downsampling from the original audio frame rate. The final output dimension matches the embedding dimension of the target LLM and fed directly into the language model with the text prompt. We evaluate the following text foundation LLMs: \textbf{Llama 3.2 1B Instruct}, \textbf{LlamA 3.1 8B Instruct}, \textbf{Llama 3.1 8B}, \textbf{DeepSeek R1 Distilled Llama 3.1 8B}, \textbf{Qwen3 0.6B, Qwen3 4B}, \textbf{Phi-4-Mini-Reasoning 3.8B}.

%These text foundation LLMs vary in two key dimensions: model scale (i.e., number of parameters) and reasoning ability (i.e., whether the model has been optimized or distilled for reasoning). This setup allows us to isolate and analyze how these LLM properties affect the downstream slot filling performance. 

This diverse selection of text LLMs spans a range of architectural scales and training specializations, including a base model, instruction-following variants, reasoning models, and hybrid configurations designed to balance instruction-following with reasoning. Our selection covers models of varying size classes (based on our size categorization), ranging from tiny (0.6-1B) to small (3–4B) and medium (8B) avoiding large and very-large LLMs for the efficiency of our experiments under limited compute resources. By choosing models from different families and training regimes, we aim to systematically investigate how model scale and capabilities influence the downstream slot-filling performance for regular (non-reasoning) and reasoning speechLLM setups. Each model was first used as the language backbone in a regular speechLLM and then incorporated into a reasoning speechLLM through supervised COT fine-tuning. This comparative approach allows us to isolate and interpret the contributions of instruction-following, reasoning ability, and model capacity to SLU performance, yielding insights into when and how reasoning capabilities translate to notable gains in SLU tasks.
\begin{table*}[t]
\small
\centering
\vspace{-10pt}
\caption{ Comparison of slot-filling performance between regular and reasoning speechLLMs.}
    \begin{tabular}{llll} 
        \hline
        \textbf{Text Foundation LLM} & \textbf{Regular SpeechLLM} & \textbf{Reasoning SpeechLLM} &\textbf{Relative Gain, $\%$} \\
                & Precision/Recall/F1  & Precision/Recall/F1 & \textbf{$\Delta$F1} \\
        \hline
          Llama 3.1 8B Instruct & {0.6292}/ 0.8726/ 0.7312  &  0.6431/ \textbf{0.9319}/ 0.7610 & +4.08 \\
        Llama 3.1 8B Base &   0.5596/ 0.9073/ 0.6923  &  \textbf{0.6691}/ 0.9168/ \textbf{0.7736} & +11.74\\
        Llama 3.2 1B Instruct &  0.5571/ 0.8541/ 0.6743 &  0.5580/ 0.9156/ 0.6934 & +2.83  \\
        Deepseek R1 Distill Llama 3.1 8B  & 0.4296/ 0.8257/ 0.5652  & 0.5616/ 0.9065/ 0.6936 & +22.72\\
        Phi4-mini reasoning 3.68B & 0.5359/ 0.8685/ 0.6628 & 0.4957/ 0.8431/ 0.6243 & -5.81 \\
        Qwen3 4B (hybrid) &  \textbf{0.6308}/ \textbf{0.9400}/ \textbf{0.7550} & 0.4979/ 0.8717/ 0.6338 & -16.05
         \\        Qwen3 0.6B (hybrid) & 0.5176/ 0.8633/ 0.6472 & 0.4889/ 0.7935/ 0.6050 &  -6.52 \\

            \hline
    \end{tabular}
 \label{tab:baselines}
\vspace{-10pt}
\end{table*}
\subsection{Regular SpeechLLMs}
All text-based LLMs described above are integrated into speechLLMs as their text foundation models. They are fine-tuned using LoRA. The modality adapter is fully fine-tuned while keeping the speech encoder, from the Whisper-base model, frozen. Here, the fine-tuning is performed using the regular (non-reasoning) slot-filling dataset with speech-prompt-response triplets as described in Section~\ref{sec:instruction}. The response is a structured JSON object containing slot types and their corresponding values without any explicit multi-step reasoning traces. These results serve as our baseline prior to introducing COT supervision in the reasoning setup which we cover in the following section. Table~\ref{tab:baselines}, "Regular SpeechLLM" column, summarizes performance of each regular speechLLM configuration determined by its text foundation LLM. We present results using partial-match precision, recall, and F1 scores. We adopt partial matching over exact match to better reflect the generative nature of the models, which may produce correct slot values with minor surface-level variations.  

Our results in this regular (non-reasoning) slot-filling setup reveal several important trends concerning model scale, specialization, training strategy and its family-specific characteristics. First, within the Llama family, we observe a clear performance change with respect to model size and instruction following abilities. The medium sized instruction-following Llama 3.1 8B-Instruct model performs the best in the family. Its base variant, Llama 3.1 8B, while not as competitive, performs surprisingly close suggesting that capacity alone,  without instruction tuning, contributes reasonably to the performance. On the other hand, the tiny instruction-tuned model, Llama 3.2 1B-Instruct, despite being a distilled version of the medium model, Llama 3.1 8B-Instruct,  performs the worst in the Llama family. This suggests that distillation, while preserving some general capabilities, may lose general language knowledge for language understanding crucial for slot filling, especially with a smaller model size. Interestingly, the reasoning variant of Llama 3.1 8B (base), distilled from a high-performing large reasoning model (Deepseek R1), Deepseek R1 Distill Llama 3.1 8B, performs worse than all other family members, with a notable performance drop. This subpar performance is likely due to the model being over-specialized for abstract reasoning tasks such as math, logic and code, which may cause it to distort linguistic knowledge and language understanding capabilities required for SLU. This observation reinforces the idea that reasoning-optimized LLMs do not guarantee performance benefits in tasks that demand general language comprehension rather than symbolic manipulation. Similarly, in a recently published study \cite{b29} it is shown that instruction-following abilities are significantly deteriorated when the models are fine-tuned for reasoning. 

In contrast, we find that the tiny hybrid model, Qwen3 0.6B, and small hybrid model, Qwen3 4B, which are trained for both reasoning and instruction-following across diverse domains, perform surprisingly well despite their relatively smaller size when compared to 8B models from the Llama series. Qwen3 4B achieved the best performance in regular speechLLMs. Their ability to balance both specializations (instruction-following and reasoning) likely lead to better generalization for the regular slot-filling task that demands some degree of implicit semantic interpretation for the direct generation of structured outputs from spoken language inputs. Notably, these hybrid models outperform the much larger reasoning-specialized model from the Llama family, Deepseek R1 Distill Llama 3.1 8B, indicating that balanced instruction-following and reasoning specialization can be more beneficial than the model size.
We also evaluated a small-size reasoning model, Phi4-mini reasoning 3.68B, from a different LLM family,  Phi4, that is open-sourced by Microsoft and optimized for a math domain, but not heavily tuned for logic and coding domains. This model performs slightly better than the tiny hybrid model, Qwen3 0.6B, likely due to its larger size, despite being primarily specialized for reasoning tasks. It also outperforms the much larger reasoning-specialized model, Deepseek R1 Distill Llama 3.1 8B, which appears to suffer from domain overfitting since it has been fine-tuned on a broader range of domains including code and logic, compared to Phi4 model's narrower focus on math. However, a similarly sized hybrid model, Qwen3 4B, significantly outperforms it, likely because it retains stronger instruction-following capabilities due to its hybrid training strategy. 

Overall, these results indicate that text foundation LLMs which are specialized solely for reasoning do not offer advantages in regular speechLLMs for slot-filling tasks that do not explicitly employ multi-step reasoning. These results also indicate the importance of model scale, balanced training objectives, and domain relevance when adapting LLMs to speech-based language understanding tasks. Notably, hybrid models trained on both reasoning and instruction data, even at smaller scales, appear to strike an effective balance between capability and generalization, making them particularly well-suited for regular SLU applications.

\begin{table*}[h]
\centering
\caption{Performance improvements achieved by hybrid speechLLMs, fine-tuned with hybrid supervision.  Results are compared against previous smaller scale models (replicated here, see Table~\ref{tab:baselines}).}
\resizebox{\textwidth}{!}{%
\begin{tabular}{llllll}
\hline

\textbf{Text Foundation Model} & \textbf{Mode} & \textbf{Regular SpeechLLM} & \textbf{Reasoning SpeechLLM} & \textbf{Hybrid SpeechLLM} & \textbf{Relative Gain, $\%$} \\
    & & Precision/Recall/F1  & Precision/Recall/F1 &  Precision/Recall/F1 & \textbf{$\Delta$F1} \\
\hline
Qwen3  0.6B       & Regular     & 0.5176/ 0.8633/ 0.6472     & -     &  0.5600/ 0.8721/ 0.6821 & +5.39\\
                  & Reasoning     &  -      & 0.4889/ 0.7935/ 0.6050    & 0.5797/ 0.8700/ 0.6958 & +15.01\\

  Qwen3  4B      & Regular       &  0.6308/ 0.9400/ 0.7550    & - & 0.6821/ 0.9340/ 0.7884 & +4.42\\
                  & Reasoning         &  -      &  0.4979/ 0.8717/ 0.6338       & 0.6958/ 0.9377/ 0.7988 & +26.03 \\
\hline
\end{tabular}
}
\vspace{-10pt}
\label{tab:hybrid}
\end{table*}
\subsection{Reasoning SpeechLLMs}
In this section we consider reasoning speechLLMs, where the models are fine-tuned using multi-step, COT style supervision for the slot-filling task. This setup enables us to assess how various text foundation LLMs differing in scale, instruction-following ability, and reasoning specialization respond to the reasoning supervision of speechLLMs. Table~\ref{tab:baselines}, "Reasoning SpeechLLM" column,  presents the results.  

We observe that the base, instruction-tuned, and 8B-parameters text foundation LLMs in reasoning speechLLMs demonstrate improvements over non-reasoning speechLLMs. However, the degree of improvement varies. Instruction-tuned foundation models show only moderate gains, suggesting that their existing alignment with instruction prompting and direct non-reasoning response generation may introduce some resistance to integrating high performant explicit multi-step reasoning traces. In contrast, the base model, which lacks prior specialization, exhibits notable improvements, and performs the best, indicating that models without post training may serve as more flexible candidates for incorporating reasoning capabilities in speechLLMs. 

The most striking observation comes from the reasoning-specialized text foundation LLM, Deepseek R1 Distill Llama 3.1 8B, which has previously shown the worst performance in the regular slot-filling setup. When its corresponding speechLLM is trained with reasoning supervision, it exhibits the largest relative performance gain of 22.72$\%$ among all other configurations. This suggests that its reasoning-focused distillation is effective when reasoning supervison becomes explicitly part of speechLLM fine-tuning. Nevertheless, this model still lags behind other models of comparable size in absolute performance, particularly when it is compared to the base model that it was distilled from. This demonstrates the cost of overfitting to a few domains (math, logic, code) and potential degradation of general linguistic knowledge and understanding during reasoning specialization.

Interestingly, smaller-scale reasoning text foundation LLMs (tiny and small) not only fail to provide benefits to the reasoning supervision of speechLLMs, but in fact cause performance degradation compared to regular speechLLMs. Despite being the top-performing model in the regular slot-filling setup, the hybrid model Qwen3 4B shows the most significant relative performance drop, approximately 16\%, when adapted to the reasoning speechLLM. This suggests that while hybrid LLMs are effective for the regular task, benefiting from implicit semantic interpretation that is likely improved due to the reasoning training of LLMs, the introduction of explicit reasoning overwhelms speechLLMs with these smaller-scale LLMs. This could be due not only to limited capacity but also to potential overfitting to the reasoning traces or a lack of regularization. In the next section, we demonstrate how hybrid training of speechLLMs helps mitigate this issue and improves overall performance.

\subsection{Hybrid SpeechLLMs}
 
To explore the potential of combining regular and reasoning response generations as two different modes of operation in a single speechLLM, we conduct experiments using hybrid fine-tuning. Here, the speechLLM is trained on a merged dataset consisting of both direct and multi-step reasoning responses. The operation is switched between regular and reasoning modes using special tags in the prompt as $\backslash$no\_think or $\backslash$think, respectively. We focus on tiny and small models from the hybrid LLM family, which previously demonstrated strong performance when compared to reasoning-only text foundation LLMs. The results are summarized in Table~\ref{tab:hybrid}. Our results show that the best-performing system is the hybrid speechLLM built on top of a hybrid text foundation LLM. 

\section{Conclusions}
In this paper, we have introduced a novel formulation of slot filling as a reasoning task and automatically created datasets supporting both regular and reasoning-style supervision. We fine-tuned speechLLMs using direct, reasoning-style, or hybrid supervision, enabling analyses across a diverse set of text foundation LLMs with varying types, sizes and prior abilities. Our findings highlight that speechLLMs with medium-scale text foundation LLMs can benefit most from reasoning supervision, while they struggle with smaller and tiny models due to lack of sufficient capacity. In general, we observed that reasoning-optimized text foundation models, focused on a single or few domains (math, logic, code), may have some loss of general language knowledge and comprehension. This degradation limits the performance of speechLLMs for slot filling, where general linguistic understanding is essential. However, we further demonstrated that hybrid fine-tuning by hybrid supervision has consistently improved the performance when using hybrid text foundation LLMs, suggesting that the hybrid training provides an effective alternative strategy for balanced fine-tuning that improves the generalization and flexibility of the models.

\end{document}